\documentclass[runningheads]{llncs}
\usepackage[T1]{fontenc}
\usepackage{graphicx}
\usepackage{booktabs}
\usepackage{multirow}
\usepackage{rotating}
\usepackage{array}
\usepackage{makecell}
\usepackage{verbatim}

\usepackage[font=small]{caption}

\begin{document}
\title{Towards Intent-Aware Human-Robot Teaming:\\ A Platform for Search-and-Rescue Operations}
\titlerunning{Towards Intent-Aware Human-Robot Teaming}
%
\author{Rohith Prem Maben\inst{1,2}\orcidID{0009-0008-9411-0695} \and
Ayesha Jena\inst{2}\orcidID{0000-0003-1735-4414} \and 
Bj{\"o}rn Olofsson\inst{3}\orcidID{0000-0003-1320-032X} \and Stefan Reitmann\inst{2,4}\orcidID{0000-0003-0283-8272} \and 
Jacek Malec\inst{2}\orcidID{0000-0002-2121-1937} \and
Rogier Woltjer \inst{1}\orcidID{0000-0001-8549-6516} \and
Elin Anna Topp\inst{2}\orcidID{0000-0001-8356-4239}} 
\authorrunning{R.P. Maben et al.}
%

\institute{%
$^{1}$ Dept. of Aviation and Aeronautical Sciences, Lund Univ., Sweden %
$^{2}$ Dept. of Computer Science, Lund Univ., Sweden, %
$^{3}$ Dept. of Automatic Control, Lund Univ., Sweden; %
$^{4}$ Chemnitz Univ. of Technology, Germany%
\\
\email{rohith.prem\_maben@aviation.lth.se,ayesha.jena@cs.lth.se}
}

%


\maketitle              
\begin{abstract}

We investigate the challenges of enabling effective collaboration between human operators and heterogeneous autonomous agents in complex, dynamic environments by developing an interaction platform that allows study of operator behavior and supports intent inference and decision-making using state-of-the-art frameworks. We demonstrate the extent to which the operator's perception, decisions, and actions could be supported by autonomous systems during search-and-rescue operations with our platform.

\keywords{Human-Robot Teaming  \and Human-Robot Interaction \and Operator Intent}
\end{abstract}

\section{Introduction}
\vspace{-0.5em}
With the increasing deployment of autonomous systems, the domain of Search and Rescue (SAR) has benefited from robots that support human operators by reducing the risks operators face and by increasing operational efficiency. Along these lines, coordinated teams of Unmanned Aerial Vehicles (UAVs) and Unmanned Ground Vehicles (UGVs) are used to locate missing persons and deliver critical supplies. A possible approach is to combine surveillance in the air with ground-level support to extend operational reach and improve the situational assessment of the operators in complex and dynamic environments~\cite{andersson2021wara}. 

However, heterogeneous teams of devices providing such a combination lead to increased challenges for the operators, arising, e.g., in maintaining balanced oversight and control of multiple agents while operating in dynamic, uncertain environments~\cite{natarajan2023human,woltjer2026designing}. Research in the field has explored efficient collaboration between operators and multi-agent teams, highlighting the need to reduce cognitive load and provide decision support~\cite{jo2024cognitive}. One way to provide such support is to examine the operator's decision-making process and to understand the intent behind these decisions. 

Previous research has investigated ways in which intent inference can be used to anticipate operator decisions, adapt system behavior, and provide timely support using probabilistic and learning-based approaches~\cite{hoffman2024inferring}. However, the application of intent inference in the context of human-robot teaming for SAR operations remains limited. There is a limited understanding of how operator intent can be effectively inferred and used to adapt control and coordination across heterogeneous multi-agent systems.

We present an interactive platform that leverages a cognitive systems model
~\cite{hollnagel2005joint} to capture operator intent through controlled SAR mission scenarios. The platform integrates three key capabilities: 1) a data collection infrastructure capturing operator interactions and system responses, 2) mission scenario configuration via Unity~\cite{unity}
with varying autonomy levels for UAV-UGV coordination with ArduPilot~\cite{a2026_ardupilot},
and 3) real-time intent inference from operator interactions through the Joint Control Framework (JCF) proposed by Lundberg \textit{et al.}~\cite{lundberg2021framework}. We demonstrate the platform through a representative SAR scenario that illustrates how operator actions reflect reliance on robot autonomy versus pre-planned procedures, and how this understanding supports adaptive decision support. 
\vspace{-0.5em}
\section{Related work}
\vspace{-1em}
Human-robot teams (HRTs) face significant challenges, with operators experiencing high cognitive load, reduced efficiency, difficulties maintaining oversight, and difficulties allocating workload effectively in dynamic environments~\cite{jo2024cognitive,rebensky2022teammates}. Creating effective HRTs requires addressing these challenges by ensuring that humans and robots can understand one another and provide decision support to operators. This can be achieved through shared mental models or dynamically adaptive levels of automation that respond to task needs while ensuring that goals are met and collaboration remains effective~\cite{singh2026human}.

Previous research has investigated how operators interact with both single and multi-robot systems, addressing aspects such as extending operator support, improving task efficiency, and optimizing task allocation methods~\cite{jena2025impact,maben2025framework,ka2024systematic}. However, as missions become more complex, there is an increasing need for more integrated approaches that can scale to the dynamic requirements of heterogeneous teams~\cite{chen2025advancing}. 
This requires unified structured frameworks that support operators across varying levels of control by inferring intent and translating it into coordinated behavior across multiple robot agents.

The JCF offers a well-structured framework that distributes sensing, deciding, and acting processes across different cognitive levels and agents using Levels of Autonomy in Cognitive Control (LACC)~\cite{lundberg2021framework}. 
While some work has been done on operationalization of the JCF for multi-robot SAR operations~\cite{hammarback2024my}, its potential to support operator intent inference is not yet extensively studied. In order to 
address this gap, we implement an integrated simulation platform built on Unity incorporating operator interfaces and data logging modules, communicating with multiple robot instances to support controlled study of operator behavior in UAV-UGV SAR operations.

\section{System architecture}
\vspace{-0.01em}
\begin{figure}[h!]
    \centering
    \includegraphics[width=1\linewidth]
    {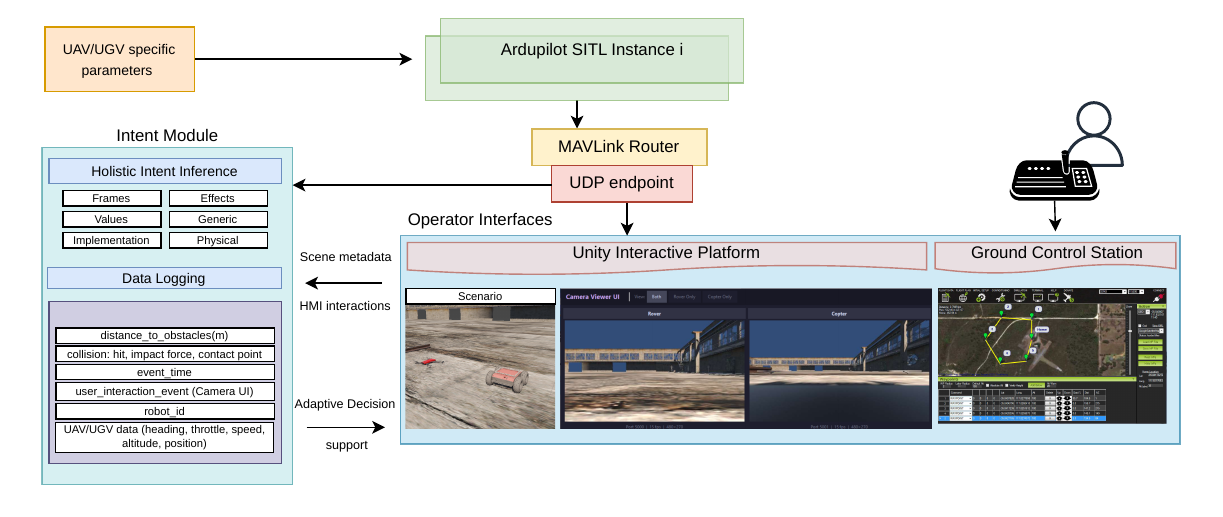}
    \caption{System architecture for Operator-UAV/UGV interaction}
    \label{fig:system_architecture}
\end{figure}

We leverage ArduPilot, a popular and actively maintained codebase supporting UGVs and UAVs, which provides a software-in-the-loop (SITL) simulator for human-machine interface investigations, with a variable-fidelity dynamics model and MAVLink ground control station compatibility. Telemetry data from one UAV and one UGV are serialized using the MAVLink v2 protocol~\cite{a2026_mavlink}
and transmitted via UDP datagrams, with four message types consumed by the Unity visualization: position, orientation, motor outputs, and flight data, as shown in Fig.~\ref{fig:system_architecture}. The MAVLink protocol ensures compatibility with standard autopilot systems such as ArduPilot, enabling the platform to interface with real-world UAV-UGV hardware without modification when transitioning from simulation to real-world deployment. A dedicated UDP port in Unity receives the UGV and UAV information and decodes incoming MAVLink frames to drive the kinematic state of the corresponding UAV-UGV model in real time. A geometrically consistent state representation is mapped from ArduPilot's North-East-Down (NED) to Unity's Left-hand coordinate scene from the first received position. 
The operator interacts with the platform via a MAVLink-enabled Ground Control Station~\cite{mission} and a Camera Viewer UI that has access to the camera streams from the UAV and the UGV in the environment. This UI enables the operator to selectively stream the UGV feed, the UAV feed, or both simultaneously via UI interaction. Additionally, a separate data acquisition module from Unity acquires relevant state information for each mission run and merges it with synchronized MAVLink telemetry into a unified per-episode log. While the current implementation presents the system with one UAV and one UGV, the architecture is designed to scale up with additional agents on dedicated UDP ports, with each agent's MAVLink telemetry mapped independently.
\footnote{https://github.com/jayesha94/IA-HRI.git}

\section{Holistic intent modeling in UAV-UGV teaming}
\vspace{-0.5em}

In order to operationalize the holistic intent structure of the operator within the SAR context and to study the interaction between humans and autonomous systems, we apply the JCF~\cite{lundberg2021framework}. 
It is intended to support modeling of temporal aspects of human-robot interaction, at diﬀerent LACC. The control levels and their operationalization in the current study context are summarized in Table \ref{tab:abstraction_hierarchy}. 
\vspace{-3em}

\begin{table}[h]
\centering
\caption{Cognitive control levels adapted for SAR human-machine interaction.}
\label{tab:abstraction_hierarchy}
\scriptsize
\renewcommand{\arraystretch}{1.2}
\setlength{\tabcolsep}{6pt}
\setlength{\arrayrulewidth}{0.4pt}
\begin{tabular}{
  >{\centering\arraybackslash}m{0.55cm}   
  >{\raggedright\arraybackslash}m{1.6cm}  
  >{\raggedright\arraybackslash}p{4.4cm}  
  >{\raggedright\arraybackslash}p{4.3cm} 
}
\toprule
\multicolumn{2}{>{\centering\arraybackslash}m{2.1cm}}{\textbf{Level}} &
\textbf{Description} &
\textbf{SAR Context} \\
\midrule
 
\multirow{2}{*}{\rotatebox{90}{\textbf{WHY}}} &
Frames &
Ascribing meaning to the situation &
\textit{Adapted mission frame} \\
 
\cmidrule(lr){2-4}
 
&
Effects &
Goals to be achieved in the ascribed situation &
Instrumental/core goal: \textit{Direct navigation, obstacle avoidance, rescue survivor} \\
 
\midrule
 
\multirow{2}{*}{\rotatebox{90}{\textbf{WHAT}}} &
Values &
Performance constraints \& trade-offs among objectives &
\textit{Navigate around obstacle, UGV safety, survivor priority} \\
 
\cmidrule(lr){2-4}
 
&
Generic &
Generic plans for common situations/Course of actions &
\textit{Avoidance maneuver, search pattern} \\
 
\midrule

\multirow{2}{*}{\rotatebox{90}{\textbf{HOW}}} &
Implementation &
UGV control actions &
\textit{Accelerating, braking, steering} \\
 
\cmidrule(lr){2-4}
 
&
Physical &
Constraints related to physical actions &
\textit{UGV commands, camera feed, terrain, survivor info} \\
 
\bottomrule
\end{tabular}
\end{table}

\vspace{-1em}

\begin{figure}[h!]
    \centering
    \includegraphics[width=11cm]{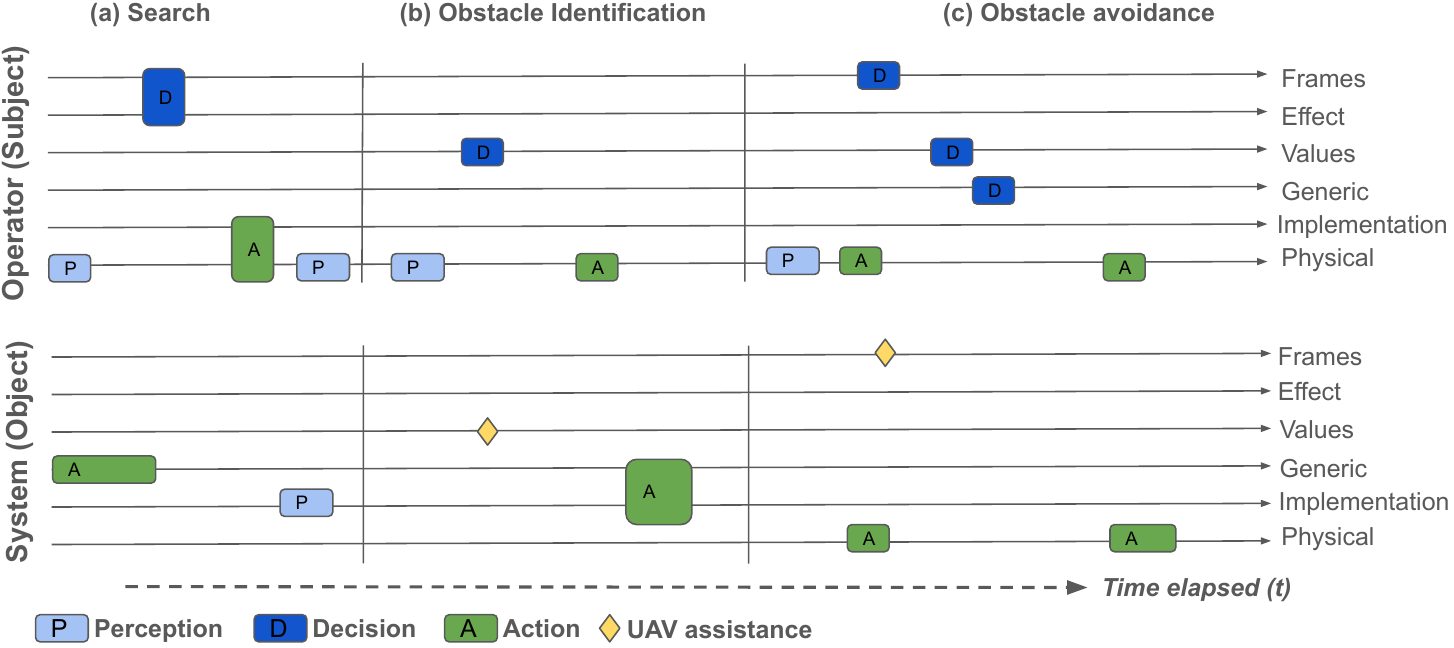}
    \caption{Human-machine interaction timeline for the scenario during an example SAR mission. The operator acts as the subject in the supervisory control process, controlling the object (UGV).}
    \label{fig:hmi_t_scenario1}
\end{figure}

We demonstrate the platform's capability for interaction analysis through Human-Machine Interaction Temporal Analysis (HMI-T)~\cite{lundberg2021framework}. As shown in Fig.~\ref{fig:hmi_t_scenario1}, the logged data decompose an example SAR scenario into perception (P), decision (D), and action (A) points across three phases: (a) search phase, (b) obstacle identification, and (c) obstacle avoidance.

\vspace{-1.5em} 
\vspace{-2em}

In phase (a), the operator begins by inspecting the environment via the UGV's raw video feed (P, Level 1), identifies a survivor by registering their map coordinates (D, Level 6/5), and initiates a slow approach while toggling heat sensors to check for additional casualties (A, Level 2/1). After assessing the path as clear, 
the operator accelerates from careful scanning to rapid transit
(D, Level 4) and accelerates along the direct line of sight (A, Level 1). In phase (b), when debris 
are detected (P, Level 1/2), 
the operator initiates a joystick retreat (A, Level 1). During phase (c), the operator then decides to handle the obstacle 
without assistance from the UAV to save time (D, Level 4/5), then commits to an alternative route (D, Level 3), and uses camera data to assess the debris size, height, and structural integrity (P, Level 2). After determining the obstacle as dangerous, the operator prioritizes stability over speed (D, Level 4) and steers the UGV safely around it (A, Level 1/2). The platform continuously logs all operator interactions throughout the scenario execution. Captured data include perception events (sensor activations, camera selection), decision points (automation levels, route commitments), and action commands (velocity adjustments, navigation). The analysis, along with data collected from the platform, reveals the function allocation between the operator and the UGV while interacting with an obstacle situation, opening up the possibility for the UAV to infer and provide adaptive assistance at the desired levels as shown in Fig.~\ref{fig:hmi_t_scenario1}.
\vspace{-0.5em}

\section{Discussion and Conclusion}
We present an interaction platform designed to evaluate holistic intent structures inference in human-autonomy teaming within SAR operations. The platform enables a systematic breakdown of a SAR operation using the JCF framework, allowing autonomous systems to infer the operator's holistic intent across multiple levels of abstraction and cognitive control. Data collection from the platform captures context-based information, including operator actions, mission context, and multi-agent interactions. This is essential for formulating intent models that reflect the dynamics of operator decision-making in complex SAR scenarios. We further define a structured approach to modeling the intention space to improve teaming through a temporal analysis of the interactions between the operator and the UGV. The approach opens up pathways towards adaptive decision support by the UAV that is in alignment with the current objectives within the operator's intent structure. This setup is representative of the broader and scalable UAV/UGV configurations that can be used. In future work, the architecture can be extended to support real-world deployment in SAR scenarios through inference, where shared intention models allow agents to provide adaptive support as required. In particular, we are working towards a Bayesian inference framework to model operator intent, using the interaction data of the platform to estimate and update probability distributions over intent space, providing an end-to-end approach for adaptive support that aligns with LACC levels.

\vspace{-0.5em}
\section{Acknowledgments}
\vspace{-0.5em}
This work was funded by ELLIIT --- the Excellence Center at Linköping University and Lund University for Information Technology, and partially supported by the Wallenberg AI, Autonomous Systems and Software Program (WASP) funded by the Knut and Alice Wallenberg Foundation. 

\bibliographystyle{splncs04}
\bibliography{bibliography}
\end{document}